\documentclass[11pt,a4paper]{article} % For LaTeX2e
\usepackage{a4wide}
\usepackage[hidelinks]{hyperref}
\usepackage{url}
\usepackage[export]{adjustbox}
\usepackage{graphicx}
\usepackage{subcaption}
\usepackage{amsmath}
\usepackage{microtype}
\usepackage{tikz}
\usepackage{booktabs}
\usepackage{natbib}

\usetikzlibrary{shapes.geometric, shadows, decorations.pathreplacing, arrows, calc, positioning, fit}

\makeatletter
\newcommand\setxveclength[5]{% newmacro, node1, anchor1, node2, anchor2
  \pgfpointdiff{\pgfpointanchor{#2}{#3}}{\pgfpointanchor{#4}{#5}}
  \edef#1{\the\pgf@x}
}
\makeatother

\title{\vspace{-2cm}Beyond Temporal Pooling: Recurrence and Temporal Convolutions for Gesture Recognition in Video}

\author{Lionel Pigou, A\"aron van den Oord\thanks{Now at Google DeepMind.} \,, Sander Dieleman\footnotemark[1]  \,,\\ Mieke Van Herreweghe \& Joni Dambre\\
\texttt{\{lionel.pigou,aaron.vandenoord,sander.dieleman,}\\
\texttt{mieke.vanherreweghe, joni.dambre\}@ugent.be} \\
Ghent University
}

\begin{document}

\maketitle

\begin{abstract}
Recent studies have demonstrated the power of recurrent neural networks for machine translation, image captioning and speech recognition. For the task of capturing temporal structure in video, however, there still remain numerous open research questions. Current research suggests using a simple temporal feature pooling strategy to take into account the temporal aspect of video. We demonstrate that this method is not sufficient for gesture recognition, where temporal information is more discriminative compared to general video classification tasks. We explore deep architectures for gesture recognition in video and propose a new end-to-end trainable neural network architecture incorporating temporal convolutions and bidirectional recurrence. Our main contributions are twofold; first, we show that recurrence is crucial for this task; second, we show that adding temporal convolutions leads to significant improvements. We evaluate the different approaches on the Montalbano gesture recognition dataset, where we achieve state-of-the-art results.
\end{abstract}

\section{Introduction} \label{sec:intro}
Gesture recognition is one of the core components in the thriving research field of human-computer interaction. 
% With computer vision techniques and/or depth sensing cameras, 
The recognition of distinct hand and arm motions is becoming increasingly important, as it enables smart interactions with electronic devices.
Furthermore, gesture identification in video can be seen as a first step towards sign language recognition, where even subtle differences in motion can play an important role. 
Some examples that complicate the identification of gestures are changes in background and lighting due to the varying environment, variations in the performance and speed of the gestures, different clothes worn by the performers and different positioning relative to the camera.
% The classification of gestures comes with a great deal of challenges. The varying environment background and lighting, different gesture performance and speed, and varying performer clothes and position relative to the camera are some examples that complicate the identification of the target gesture. 
% Moreover, some hand gestures only differ in finger position, meaning that very small pixel patches can be greatly discriminatory. 
Moreover, regular hand motion or out-of-vocabulary gestures should not to be confused with one of the target gestures.

Convolutional neural networks (CNNs) \citep{lecun1998gradient} are the de facto standard approach in computer vision. 
CNNs have the ability to learn complex hierarchies with increasing levels of abstraction while being end-to-end trainable.
% of features that prove to be highly effective in contrast to handcrafted features. 
Their success has had a huge impact on vision based applications like image classification \citep{krizhevsky2012imagenet}, object detection \citep{sermanet2013overfeat}, human pose estimation \citep{toshev2014deeppose} and many more.  
A video can be seen as an ordered collection of images. 
Classifying a video frame by frame with a CNN is bound to ignore motion characteristics, as there is no integration of temporal information.
Depending on the task at hand, aggregating the spatial features produced by the CNN with temporal pooling can be a viable strategy \citep{karpathy2014large,ng2015beyond}. As we'll show in this paper, however, this method is of limited use for gesture recognition. %We need to model gesture motion with a different architecture.

Apart from a collection of frames, a video can also be seen as a time series.
Some of the most successful models for time series classification are recurrent neural networks (RNNs) with either standard cells or long short-term memory (LSTM) cells \citep{hochreiter1997long}.
% They are so-called ``deep in time'', because
Their ability to learn dynamic temporal dependencies has allowed researchers to achieve breakthrough results in e.g. speech recognition \citep{graves2013speech}, machine translation \citep{sutskever2014sequence} and image captioning \citep{vinyals2014show}.
% Training these recurrent models directly on video would perform poorly. 
% Video tends to be spatially high dimensional, thus prone to overfitting. 
Before feeding video to recurrent models, we need to incorporate some form of spatial or spatiotemporal feature extraction. This motivates the concept of combining CNNs with RNNs. CNNs have unparalleled spatial (and spatiotemporal with added temporal convolutions) feature extraction capabilities, while adding recurrence ensures the modeling of feature evolution over time.

% So it wouldn't be the worst idea to combine CNNs for their learned spatial features and LSTMs or RNNs for their temporal features trying to learn spatiotemporal features. 

For general video classification datasets like UCF-101 \citep{soomro2012ucf101}, Sports-1M \citep{karpathy2014large} or HMDB-51 \citep{kuehne2011hmdb}, the temporal aspect is of less importance compared to a gesture recognition dataset. For example, the appearance of a violin almost certainly suggests the target class is ``playing violin'', as no other class involves a violin. 
The model has no need to capture motion information for this particular example. 
That being said, there are some categories where modeling motion in some way or another is always beneficial. In the case of gesture recognition, however, motion plays a more critical role. Many gestures are not only defined by their spatial hand and/or arm placement, but also by their motion pattern.
% Take note that our work looks at frame-wise classification, while these datasets provide labels to classify entire video files or video fragments.

In this work, we explore a variety of end-to-end trainable deep networks for video classification applied to frame-wise gesture recognition with the Montalbano dataset that was introduced in the ChaLearn LAP 2014 Challenge \citep{chalearn14}. We study two ways of capturing the temporal structure of these videos. The first method involves temporal convolutions to enable the learning of motion features. The second method introduces recurrence to  our networks, which allows the modeling of temporal dynamics, which plays an essential role in gesture recognition. 
% We show in this paper that RNNs play an essential role in gesture recognition.
% Our CNN baseline models (Figure \ref{fig:baseline}) are: (i) a single-frame architecture and (ii) a temporal feature pooling strategy. To improve the baseline, we examine (i) a CNN with added temporal convolutions and max-pooling acting on three dimensions, (ii) a CNN stacked by a bidirectional recurrent network and (iii) we propose a new architecture which consists of a CNN with added temporal convolutions and bidirectional recurrence.

% All our networks are trained end-to-end without intermediate steps.

 % For this reason, a temporal feature pooling architecture suffices to aggregate the target class, and, as shown in [r], is even more beneficial than LSTMs. 

% In this paper, we tackle the challenging problem of gesture recognition. The goal is to continuously recognize hand gestures in video data in a supervised manner (supervised frame-wise video classification). <why challenging, why important> 
% Although there are undoubtedly gestures that have unique enough properties to be recognized using a single frame, there is no telling. 
% We propose the first end-to-end trainable deep neural network that combines convolutional layers with recurrent units that outperforms temporal feature pooling for video classification.

\section{Related Work} \label{sec:relwork}

An extensive evaluation of CNNs on general video classification is provided by \citet{karpathy2014large} using the Sports-1M dataset. They compare different frame fusion methods to a baseline single-frame architecture and conclude that their best fusion strategy only modestly improves the accuracy of the baseline.
Their work is extended by \citet{ng2015beyond}, who show that LSTMs achieve no improvements over a temporal feature pooling scheme on the UCF-101 dataset for human action classification and only marginal improvements on the Sports-1M dataset. For this reason, the single-frame and the temporal pooling architectures are important baseline models.

Another way to capture motion is to convert a video stream to a dense optical flow. This is a way to represent motion spatially by estimating displacement vectors of each pixel. It is a core component in the two-stream architecture described by \cite{simonyan2014two} and is used for human pose estimation \citep{jain2014modeep}, for global video descriptor learning \citep{ng2015beyond} and for video captioning \citep{venugopalan2015sequence}. We have not experimented with optical flow, because (i) it has a greater computational preprocessing complexity and (ii) our models should implicitly learn to infer motion features in an end-to-end fashion, so we chose not to engineer them.

\cite{neverova2014moddrop} present an extended overview of their winning solution for the ChaLearn LAP 2014 gesture recognition challenge and achieve a state-of-the-art score on the Montalbano dataset. They propose a multi-modal `ModDrop' network operating at three temporal scales and use an ensemble method to merge the features at different scales. They also developed a new training strategy, ModDrop, that makes the network's predictions robust to missing or corrupted channels.
% We opted not to consider this as our baseline, because their work has a focus on the multi-modality of the data and is not end-to-end trainable, whereas our work centers on simple neural architectures that are easily trainable. 

Most of the constituent parts in our architectures have been used before in other work for different purposes. Learning motion features with three-dimensional convolution layers has been studied by \cite{ji20133d} and \cite{taylor2010convolutional} to classify short clips of human actions on the KTH dataset. \cite{baccouche2011sequential} proposed  including a  two-step scheme to model the temporal evolution of learned features with an LSTM. 
% A three-dimensional max-pooling architecture for gesture recognition is proposed by Pigou et al. \cite{pigou2014sign}. 
Finally, the combination of a CNN with an RNN has been used for speech recognition \citep{hannun2014deepspeech}, image captioning \citep{vinyals2014show} and video narration \citep{donahue2015long}.

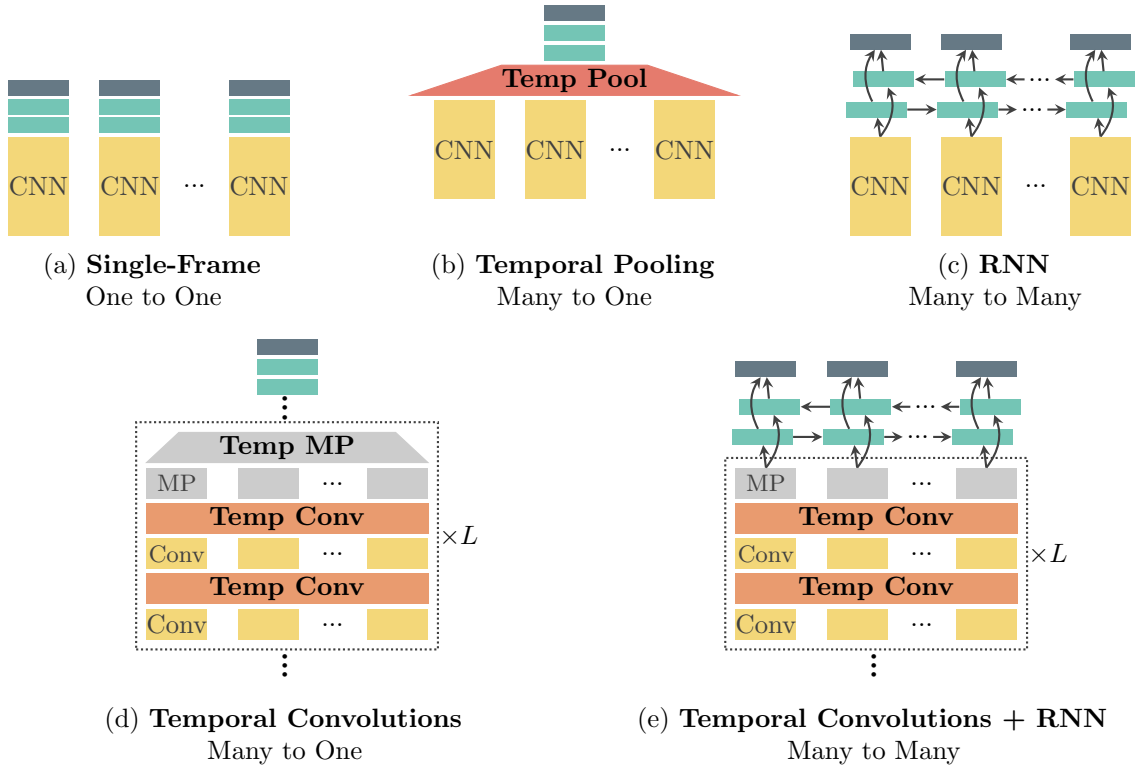
\begin{figure}[tb]
\def \rsize {3cm}
\begin{center}
\captionsetup[subfigure]{justification=centering,singlelinecheck=true}
% \hspace*{\fill}
\begin{subfigure}[b]{0.28\textwidth}
\centering
%!TEX root = ../nips2015.tex
\begin{tikzpicture}[font=\small, every node/.style={very thin}]

% rgb(51,77,92)
\definecolor{blu}{RGB}{51,77,92}
\definecolor{ye}{RGB}{239,201,76}
\definecolor{gre}{RGB}{69,178,157}
\def \cnnxshift {0.2cm}
\def \cnnxshiftn {0.7cm}
\def \arrsize {0.6cm}
\def \circsize {0.3cm}
\def \spacingz {0.5mm}
\def \cnnwidth {0.8cm}

\tikzstyle{cnn} = [
	rectangle, 
	inner sep=0pt,
	minimum width=\cnnwidth, 
	minimum height=1.3cm,
	text centered, 
	% draw=black!75,
	text=black!75,
	% text=white!100,
	fill=ye!75]

\tikzstyle{dense} = [
	outer sep=0pt,
	rectangle, 
	inner sep=0pt,
	minimum width=\cnnwidth, 
	minimum height=0.2cm,
	text centered, 
	% draw=black!75,
	text=black!75,
	fill=gre!75]

\tikzstyle{softmax} = [
	outer sep=0pt,
	rectangle, 
	inner sep=0pt,
	minimum width=\cnnwidth, 
	minimum height=0.2cm,
	text centered, 
	% draw=black!75,
	text=black!75,
	fill=blu!75]

\tikzstyle{arrow} = [thick, ->,>=stealth]

\tikzstyle{farr} = [bend right, looseness=1]
\tikzstyle{farl} = [bend left]

\node (dummy) [rectangle, minimum height=2cm] {};

\node (cnn1) [cnn, below = 0cm of dummy.south, anchor=north] {CNN};
\node (cnn2) [cnn, right of=cnn1, xshift=\cnnxshift] {CNN};
\node (cnn3) [cnn, right of=cnn2, xshift=\cnnxshiftn] {CNN};
\node (ell) at ($(cnn2)!0.5!(cnn3)$) {...};

% \node (ell) [right= 0.2cm of cnn2.east, anchor=west] {...};
% \node (ell2) [left= 0.2cm of cnn1.west, anchor=east] {...};
 
% \setxveclength{\mydist}{cnn1}{west}{cnn3}{east}
% \node (pool1) [pool, above = \spacingz of cnn1.north west, anchor=south west, minimum width=\mydist] {Temporal Pooling};

\node (d1) [dense, above = \spacingz of cnn1.north, anchor=south, text=white] {\tiny};
\node (d2) [dense, above = \spacingz of d1.north, anchor=south, text=white] {\tiny};
\node (softmax1) [softmax, above = \spacingz of d2.north, anchor=south, text=white] {\tiny};

\node (d11) [dense, above = \spacingz of cnn2.north, anchor=south] {};
\node (d21) [dense, above = \spacingz of d11.north, anchor=south] {};
\node (softmax11) [softmax, above = \spacingz of d21.north, anchor=south] {};

\node (d12) [dense, above = \spacingz of cnn3.north, anchor=south] {};
\node (d22) [dense, above = \spacingz of d12.north, anchor=south] {};
\node (softmax12) [softmax, above = \spacingz of d22.north, anchor=south] {};

\end{tikzpicture}
% \resizebox{!}{\rsize}{\input{imgs/singlen.tex}}
\caption{\textbf{Single-Frame} \\ One to One}
\label{fig:archsingle}
\end{subfigure}
\hfill
\begin{subfigure}[b]{0.28\textwidth}
\centering
%!TEX root = ../nips2015.tex
\begin{tikzpicture}[font=\small]

\def \cnnxshift {0.2cm}
\def \cnnxshiftn {0.7cm}
\def \arrsize {0.6cm}
\def \circsize {0.3cm}
\def \spacingz {0.6mm}
\def \cnnwidth {0.8cm}
\definecolor{blu}{RGB}{51,77,92}
\definecolor{ye}{RGB}{239,201,76}
\definecolor{gre}{RGB}{69,178,157}
% rgb(223,90,73)
\definecolor{redd}{RGB}{223,90,73}
% rgb(39,39,39)

\tikzstyle{cnn} = [
	rectangle, 
	inner sep=0pt,
	minimum width=\cnnwidth, 
	minimum height=1.3cm,
	text centered, 
	% draw=black!75,
	text=black!75,
	% text=white!100,
	fill=ye!75]

% mytrap/.style={
%   trapezium, trapezium angle=67.5, draw,inner xsep=0pt,outer sep=0pt,
%   minimum height=1.81mm, text width=#1
% }

\tikzstyle{pool} = [
	trapezium, trapezium angle=15, draw, inner sep=0pt,
	trapezium stretches=true,
  minimum height=0.4cm,
  text height = 0.25cm,
  minimum width=0cm,
  inner xsep=0pt,
  inner ysep=0pt,
  % text widttext widthh=\cnnwidth,
  text centered, 
	draw=redd!80,
	fill=redd!80
]

% \tikzstyle{pool} = [
% 	rectangle, 
% 	outer sep=0pt,
% 	inner sep=0pt,
% 	minimum height=0.4cm,
% 	text centered, 
% 	% draw=black,
% 	fill=redd!80]

\tikzstyle{dense} = [
	outer sep=0pt,
	rectangle, 
	inner sep=0pt,
	minimum width=\cnnwidth, 
	minimum height=0.2cm,
	text centered, 
	% draw=black!75,
	text=black!75,
	fill=gre!75]

\tikzstyle{softmax} = [
	outer sep=0pt,
	rectangle, 
	inner sep=0pt,
	minimum width=\cnnwidth, 
	minimum height=0.2cm,
	text centered, 
	% draw=black!75,
	text=black!75,
	fill=blu!75]

\tikzstyle{arrow} = [thick, ->,>=stealth]

\tikzstyle{farr} = [bend right, looseness=1]
\tikzstyle{farl} = [bend left]

\node (cnn1) [cnn] {CNN};
\node (cnn2) [cnn, right of=cnn1, xshift=\cnnxshift] {CNN};
\node (cnn3) [cnn, right of=cnn2, xshift=\cnnxshiftn] {CNN};
\node (ell) at ($(cnn2)!0.5!(cnn3)$) {...};
 
\setxveclength{\mydist}{cnn1}{west}{cnn3}{east}
\node (pool1) [pool, above = \spacingz of cnn1.north west, anchor=south west, minimum width=\mydist, xshift=0.445*\mydist] {\bf Temp Pool};

\node (dense1) [dense, above = \spacingz of pool1.north, anchor=south] {};
\node (dense2) [dense, above = \spacingz of dense1.north, anchor=south] {};
\node (softmax1) [softmax, above = \spacingz of dense2.north, anchor=south] {};

\end{tikzpicture}
% \resizebox{!}{\rsize}{\input{imgs/pool.tex}}
\caption{\textbf{Temporal Pooling} \\ Many to One}
\label{fig:archpool}
\end{subfigure} 
\hfill
\begin{subfigure}[b]{0.28\textwidth}
\centering
%!TEX root = ../nips2015.tex
\begin{tikzpicture}[font=\small]

\def \cnnxshift {0.2cm}
\def \cnnxshiftn {0.7cm}

\def \cnncellyshift {2.5mm}
\def \cellcellyshift {-0.6cm}
\def \celltopyshift {-0.5cm}

\def \arrsize {0.65cm}
\def \circsize {0.2cm}
\def \circsizen {0.8cm}
\def \cnnwidth {0.8cm}

\definecolor{blu}{RGB}{51,77,92}
\definecolor{ye}{RGB}{239,201,76}
\definecolor{gre}{RGB}{69,178,157}
% rgb(223,90,73)
\definecolor{redd}{RGB}{223,90,73}

\tikzstyle{cnn} = [
	rectangle, 
	inner sep=0pt,
	minimum width=\cnnwidth, 
	minimum height=1.3cm,
	text centered, 
	% draw=black!75,
	text=black!75,
	% text=white!100,
	fill=ye!75]

\tikzstyle{cell} = [
	rectangle, 
	minimum height=\circsize,
	minimum width=\circsizen,
	inner sep=0pt,
	text centered, 
	fill=gre!75]

\tikzstyle{cellleft} = [
	rectangle, 
	minimum height=\circsize,
	minimum width=\circsizen,
	inner sep=0pt,
	text centered, 
	fill=gre!75]

\tikzstyle{cellout} = [
	rectangle, 
	minimum height=\circsize,
	minimum width=\circsizen,
	inner sep=0pt,
	text centered, 
	fill=blu!75]

\tikzstyle{arrow} = [thick, ->,>=stealth, black!75]

\tikzstyle{farr} = [bend right, looseness=1]
\tikzstyle{farl} = [bend left]

\node (cnn1) [cnn] {CNN};
\node (cnn2) [cnn, right of=cnn1, xshift=\cnnxshift] {CNN};
\node (cnn3) [cnn, right of=cnn2, xshift=\cnnxshiftn] {CNN};

\foreach \i in {1,2,3}{
	% \node (rnnobx\i) [rnnbox, above of=cnn\i, yshift=\cnnrnnyshift, anchor=south] {};

	\node (cellb\i) [cell, above = \cnncellyshift of cnn\i, xshift=-0.05cm, yshift=0cm, anchor=south] {};

	\node (anch\i) [above = 0cm of cnn\i.north, yshift=0cm, minimum height=0pt,  minimum width=0pt, anchor=south, inner sep=0pt, outer sep=0pt] {};

	\node (cella\i) [cellleft, above of=cellb\i, xshift=0.1cm,yshift=\cellcellyshift] {};

	\node (cellt\i) [cellout, above of=cella\i, xshift=-0.05cm, yshift=\celltopyshift] {};

	\draw [arrow] (anch\i) -- (cellb\i);
	\draw [arrow] (anch\i) edge[farr] (cella\i);
	\draw [arrow] (cellb\i) edge[farl] (cellt\i);
	\draw [arrow] (cella\i) -- (cellt\i);
	% \draw [arrow] (cnn\i) -- (rnnbox\i);
	% \draw [arrow] (rnnbox\i) -- (cellt\i);
}

\draw [arrow] (cellb1) -- (cellb2);
\draw [arrow] (cella2) -- (cella1);

\draw [arrow] (cellb2) -- ++(\arrsize,0);
\draw [arrow, <-] (cella2) -- ++(\arrsize,0);
\draw [arrow] (cella3) -- ++(-\arrsize,0);
\draw [arrow, <-] (cellb3) -- ++(-\arrsize,0);
% \draw [arrow] (cella2) -- (cella1);
% \draw [arrow] (cella2) -- (cella1);
% \draw [arrow] (cella2) -- (cella1);

% \node (elltop) at ($(rnnbox2)!0.5!(rnnbox3)$) {...};

\node (ell1) at ($(cnn2)!0.5!(cnn3)$) {...};
\node (ell2) at ($(cella2)!0.5!(cella3)$) {...};
\node (ell3) at ($(cellb2)!0.5!(cellb3)$) {...};

\end{tikzpicture}
% \resizebox{!}{\rsize}{\input{imgs/arch2.tex}}
\caption{ \textbf{RNN} \\ Many to Many}
\label{fig:archrec}
\end{subfigure}
% \hspace*{\fill}
\vspace{0.3cm}

\begin{subfigure}[b]{0.49\textwidth}
\centering
%!TEX root = ../nips2015.tex
\begin{tikzpicture}[font=\small]

\def \cnnxshift {0.2cm}
\def \cnnxshiftn {0.7cm}
\def \arrsize {0.6cm}
\def \circsize {0.3cm}
\def \spacingz {0.6mm}
\def \cnnwidth {0.8cm}
\definecolor{blu}{RGB}{51,77,92}
\definecolor{ye}{RGB}{239,201,76}
\definecolor{gre}{RGB}{69,178,157}
% rgb(223,90,73)
\definecolor{ora}{RGB}{226,122,63}
\definecolor{redd}{RGB}{223,90,73}

\tikzstyle{2dconv} = [
	rectangle, 
	inner sep=0pt,
	minimum width=\cnnwidth, 
	minimum height=0.4cm,
	text centered, 
	% draw=black!100,
	text=black!75,
	fill=ye!75]

\tikzstyle{1dconv} = [
	outer sep=0pt,
	rectangle, 
	inner sep=0pt,
	minimum height=0.4cm,
	text centered, 
	% draw=black!100,
	% text=black!100,
	fill=ora!75]

% \tikzstyle{3dmp} = [
% 	outer sep=0pt,
% 	rectangle, 
% 	inner sep=0pt,
% 	minimum height=0.4cm,
% 	text centered, 
% 	% draw=black!100,
% 	% text=black!100,
% 	fill=blue!30]

\tikzstyle{3dmp} = [
	trapezium, trapezium angle=50, draw, inner sep=0pt,
	trapezium stretches=true,
  minimum height=0.4cm,
  text height = 0.25cm,
  minimum width=0cm,
  inner xsep=0pt,
  inner ysep=0pt,
  % text widttext widthh=\cnnwidth,
  text centered, 
	draw=black!20,
	fill=black!20
]

\tikzstyle{2dmp} = [
	rectangle, 
	inner sep=0pt,
	minimum width=\cnnwidth, 
	minimum height=0.4cm,
	text centered, 
	% draw=black!100,
	text=black!75,
	fill=black!20]

\tikzstyle{dense} = [
	outer sep=0pt,
	rectangle, 
	inner sep=0pt,
	minimum width=\cnnwidth, 
	minimum height=0.2cm,
	text centered, 
	% draw=black!75,
	text=black!75,
	fill=gre!75]

\tikzstyle{softmax} = [
	outer sep=0pt,
	rectangle, 
	inner sep=0pt,
	minimum width=\cnnwidth, 
	minimum height=0.2cm,
	text centered, 
	% draw=black!75,
	text=black!75,
	fill=blu!75]

\tikzstyle{arrow} = [thick, ->,>=stealth]

\tikzstyle{farr} = [bend right, looseness=1]
\tikzstyle{farl} = [bend left]

\node (2d1) [2dconv] {Conv};
\node (2d2) [2dconv, right of=2d1, xshift=\cnnxshift] {};
\node (2d3) [2dconv, right of=2d2, xshift=\cnnxshiftn] {};
\node (ell) at ($(2d2)!0.5!(2d3)$) {...};

 \setxveclength{\mydist}{2d1}{west}{2d3}{east}
\node (1d1) [1dconv, above = \spacingz of 2d1.north west, anchor=south west, minimum width=\mydist] {\bf Temp Conv};

\node (2d11) [2dconv, above = \spacingz of 1d1.north west, anchor=south west] {\footnotesize Conv};
\node (2d21) [2dconv, right of=2d11, xshift=\cnnxshift] {};
\node (2d31) [2dconv, right of=2d21, xshift=\cnnxshiftn] {};
\node (ell) at ($(2d21)!0.5!(2d31)$) {...};

 \setxveclength{\mydist}{2d11}{west}{2d31}{east}
\node (1d11) [1dconv, above = \spacingz of 2d11.north west, anchor=south west, minimum width=\mydist] {\bf Temp Conv};

\node (2dm1) [2dmp, above = \spacingz of 1d11.north west, anchor=south west] {\footnotesize MP};
\node (2dm2) [2dmp, right of=2dm1, xshift=\cnnxshift] {};
\node (2dm3) [2dmp, right of=2dm2, xshift=\cnnxshiftn] {};
\node (ell) at ($(2dm2)!0.5!(2dm3)$) {...};

\node (mp1) [3dmp, above = \spacingz of 2dm1.north west, anchor=south west, minimum width=\mydist, xshift=0.445*\mydist] {\bf Temp MP};

\node (boxr) [draw=black!75, densely dotted,thick,fit=(mp1) (2d1) (2d3)] {};

% \node (2d11) [2dconv, above = \spacingz of mp1.north, anchor=south, xshift=-1cm] {2D};
% \node (2d21) [2dconv, right of=2d11, xshift=0.1*\cnnxshift] {2D};
% \node (2d31) [2dconv, right of=2d21, xshift=0.1*\cnnxshiftn] {2D};
% \node (ell) at ($(2d21)!0.5!(2d31)$) {...};

%  \setxveclength{\mydist}{2d11}{west}{2d31}{east}
% \node (1d11) [1dconv, above = \spacingz of 2d11.north west, anchor=south west, minimum width=\mydist] {\footnotesize Temp Conv};

% \node (mp11) [3dmp, above = \spacingz of 1d11.north, anchor=south, minimum width=\mydist] {\footnotesize 3D Max-Pool};

\node(d1) [dense, above= 0.5cm of mp1.north, anchor=south] {};
\node(d2) [dense, above= \spacingz of d1.north, anchor=south] {};
\node(s2) [softmax, above= \spacingz of d2.north, anchor=south] {};

\node (ell1) [below= 0.2cm of boxr.south, anchor=base, rotate=90] {\normalsize \textbf{...}};
\node (ell2) at ($(boxr.north)!0.5!(d1.south)$) [anchor=base, rotate=90] {\normalsize \textbf{...}};

% \node (ell2) [above= 0.2cm of boxr.north, anchor=base, rotate=90] {...};

\node (maalk) [right = 0cm of boxr.east, inner sep=0] {$\times$$L$};
\node (maalk2) [left = 0cm of boxr.west, inner sep=0, anchor=east, minimum width=0.7cm] {};

 % \setxveclength{\mydist}{mp1}{north east}{ell}{south east}
% \draw [decorate, decoration={brace, raise=0.2mm, amplitude=1.5mm, mirror}, thick] (mp1.north west) -- ++(0,-1.75cm) 
% 	node (L) [black,midway,xshift=-1.5mm, anchor=east] {$\times 4$};

% \draw [decorate, decoration={brace, raise=0.2mm, amplitude=1.5mm}, thick] (1d1.north east) -- (2d3.south east) 
% 	node (L) [black,midway,xshift=0.6mm, anchor=west] {$\times 2$};

% \draw [decorate, decoration={brace, raise=7mm, amplitude=1.5mm,}, thick] (mp1.north east) -- ++(0,-1.75cm) 
% 	node (L) [black,midway,xshift=7.6mm, anchor=west] {$\times 4$};

% \draw [decorate, decoration={brace, raise=0.4mm, amplitude=1.5mm}, thick] (1d1.north east) -- ++(0,-1.3cm) 
% 	node (L) [black,midway,xshift=0.6mm, anchor=west] {$\times 2$};

% \draw [arrow] (L) edge[farl, out=90] (ell);

% \node (cnn1) [cnn] {CNN};
% \node (cnn2) [cnn, right of=cnn1, xshift=\cnnxshift] {CNN};
% \node (cnn3) [cnn, right of=cnn2, xshift=\cnnxshiftn] {CNN};
% \node (ell) at ($(cnn2)!0.5!(cnn3)$) {...};
 
% \setxveclength{\mydist}{cnn1}{west}{cnn3}{east}
% \node (pool1) [pool, above = \spacingz of cnn1.north west, anchor=south west, minimum width=\mydist] {Temporal Pooling};

% \node (dense1) [dense, above = \spacingz of pool1.north, anchor=south] {Dense};
% \node (dense2) [dense, above = \spacingz of dense1.north, anchor=south] {Dense};
% \node (softmax1) [softmax, above = \spacingz of dense2.north, anchor=south] {Softmax};

\end{tikzpicture}
% \resizebox{!}{\rsize}{\input{imgs/conv3d.tex}}
\caption{\textbf{Temporal Convolutions} \\ Many to One}
\label{fig:arch3d}
\end{subfigure}
% \vspace{0.3cm}
%
\begin{subfigure}[b]{0.49\textwidth}
\centering
%!TEX root = ../nips2015.tex
\begin{tikzpicture}[font=\small]

\def \cnnxshift {0.2cm}
\def \cnnxshiftn {0.7cm}
\def \arrsize {0.6cm}
\def \circsize {0.3cm}
\def \spacingz {0.6mm}
\def \cnnwidth {0.8cm}

\def \cnncellyshift {3mm}
\def \cellcellyshift {-0.6cm}
\def \celltopyshift {-0.5cm}

\def \arrsize {0.65cm}
\def \circsize {0.2cm}
\def \circsizen {0.8cm}
\def \cnnwidth {0.8cm}

\definecolor{blu}{RGB}{51,77,92}
\definecolor{ye}{RGB}{239,201,76}
\definecolor{gre}{RGB}{69,178,157}
% rgb(223,90,73)
\definecolor{ora}{RGB}{226,122,63}
\definecolor{redd}{RGB}{223,90,73}

\tikzstyle{2dconv} = [
	rectangle, 
	inner sep=0pt,
	minimum width=\cnnwidth, 
	minimum height=0.4cm,
	text centered, 
	% draw=black!100,
	text=black!75,
	fill=ye!75]

\tikzstyle{1dconv} = [
	outer sep=0pt,
	rectangle, 
	inner sep=0pt,
	minimum height=0.4cm,
	text centered, 
	% draw=black!100,
	% text=black!100,
	fill=ora!75]

\tikzstyle{3dmp} = [
	outer sep=0pt,
	rectangle, 
	inner sep=0pt,
	minimum height=0.3cm,
	text centered, 
	% draw=black!100,
	% text=black!100,
	fill=black!20]

\tikzstyle{2dmp} = [
	rectangle, 
	inner sep=0pt,
	minimum width=\cnnwidth, 
	minimum height=0.4cm,
	text centered, 
	% draw=black!100,
	text=black!75,
	fill=black!20]

\tikzstyle{cell} = [
	rectangle, 
	minimum height=\circsize,
	minimum width=\circsizen,
	inner sep=0pt,
	text centered, 
	fill=gre!75]

\tikzstyle{cellleft} = [
	rectangle, 
	minimum height=\circsize,
	minimum width=\circsizen,
	inner sep=0pt,
	text centered, 
	fill=gre!75]

\tikzstyle{cellout} = [
	rectangle, 
	minimum height=\circsize,
	minimum width=\circsizen,
	inner sep=0pt,
	text centered, 
	fill=blu!75]

\tikzstyle{arrow} = [thick, ->,>=stealth, black!75]

\tikzstyle{farr} = [bend right, looseness=1]
\tikzstyle{farl} = [bend left]

\node (2d1) [2dconv] {Conv};
\node (2d2) [2dconv, right of=2d1, xshift=\cnnxshift] {};
\node (2d3) [2dconv, right of=2d2, xshift=\cnnxshiftn] {};
\node (ell) at ($(2d2)!0.5!(2d3)$) {...};

 \setxveclength{\mydist}{2d1}{west}{2d3}{east}
\node (1d1) [1dconv, above = \spacingz of 2d1.north west, anchor=south west, minimum width=\mydist] {\bf Temp Conv};

\node (2d11) [2dconv, above = \spacingz of 1d1.north west, anchor=south west] {\footnotesize Conv};
\node (2d21) [2dconv, right of=2d11, xshift=\cnnxshift] {};
\node (2d31) [2dconv, right of=2d21, xshift=\cnnxshiftn] {};
\node (ell) at ($(2d21)!0.5!(2d31)$) {...};

 \setxveclength{\mydist}{2d11}{west}{2d31}{east}
\node (1d11) [1dconv, above = \spacingz of 2d11.north west, anchor=south west, minimum width=\mydist] {\bf Temp Conv};

\node (2dm1) [2dmp, above = \spacingz of 1d11.north west, anchor=south west] {\footnotesize MP};
\node (2dm2) [2dmp, right of=2dm1, xshift=\cnnxshift] {};
\node (2dm3) [2dmp, right of=2dm2, xshift=\cnnxshiftn] {};
\node (ell) at ($(2dm2)!0.5!(2dm3)$) {...};

% \node (mp1) [3dmp, above = \spacingz of 2dm1.north west, anchor=south west, minimum width=\mydist] {\scriptsize Temporal Max-Pooling};

\node (boxr) [draw=black!75, densely dotted, thick,fit=(2dm1) (2dm3) (2d2) (2d3)] {};

% \node (ell1) [below= 0.2cm of boxr.south, anchor=base, rotate=90] {...};
% \node (ell2) at ($(boxr.north)!0.5!(d1.south)$) [anchor=base, rotate=90] {...};

\foreach \i in {1,2,3}{
	% \node (rnnobx\i) [rnnbox, above of=cnn\i, yshift=\cnnrnnyshift, anchor=south] {};

	\node (cellb\i) [cell, above =\cnncellyshift of 2dm\i, xshift=-0.05cm, yshift=0cm, anchor=south] {};

	\node (anch\i) [above = 0cm of 2dm\i.north, yshift=0cm, minimum height=0pt,  minimum width=0pt, anchor=south, inner sep=0pt, outer sep=0pt] {};

	\node (cella\i) [cellleft, above of=cellb\i, xshift=0.1cm,yshift=\cellcellyshift] {};

	\node (cellt\i) [cellout, above of=cella\i, xshift=-0.05cm, yshift=\celltopyshift] {};

	\draw [arrow] (anch\i) -- (cellb\i);
	\draw [arrow] (anch\i) edge[farr] (cella\i);
	\draw [arrow] (cellb\i) edge[farl] (cellt\i);
	\draw [arrow] (cella\i) -- (cellt\i);
	% \draw [arrow] (cnn\i) -- (rnnbox\i);
	% \draw [arrow] (rnnbox\i) -- (cellt\i);
}

\draw [arrow] (cellb1) -- (cellb2);
\draw [arrow] (cella2) -- (cella1);

\draw [arrow] (cellb2) -- ++(\arrsize,0);
\draw [arrow, <-] (cella2) -- ++(\arrsize,0);
\draw [arrow] (cella3) -- ++(-\arrsize,0);
\draw [arrow, <-] (cellb3) -- ++(-\arrsize,0);
% \draw [arrow] (cella2) -- (cella1);
% \draw [arrow] (cella2) -- (cella1);
% \draw [arrow] (cella2) -- (cella1);

% \node (elltop) at ($(rnnbox2)!0.5!(rnnbox3)$) {...};

% \node (ell1) at ($(cnn2)!0.5!(cnn3)$) {...};
\node (ell2) at ($(cella2)!0.5!(cella3)$) {...};
\node (ell3) at ($(cellb2)!0.5!(cellb3)$) {...};
\node (maalk) [right = 0cm of boxr.east, inner sep=0] {$\times$$L$};
\node (maalk2) [left = 0cm of boxr.west, inner sep=0, anchor=east, minimum width=0.7cm] {};

\node (ell1) [below= 0.2cm of boxr.south, anchor=base, rotate=90] {\normalsize \textbf{...}};

\end{tikzpicture}
% \resizebox{!}{\rsize}{\input{imgs/tconvrnn.tex}}
% \input{imgs/tconvrnn.tex}
\caption{\textbf{Temporal Convolutions + RNN} \\ Many to Many}
\label{fig:arch3drec}
\end{subfigure}
\end{center}
\caption{\textbf{Overview}\,  (a) Single-frame CNN architecture. (b) Temporal feature pooling network (max- or mean-pooling), spanning multiple video frames. (c) Model with bidirectional recurrence. (d) Adding temporal convolutions and three-dimensional max-pooling (MP refers to max-pooling). (e) Architecture with added temporal convolutions and bidirectional recurrence.}
% Panel (a) shows a single-frame CNN architecture. Panel (b) illustrates a temporal feature pooling network (max- or mean-pooling), spanning multiple video frames. In (c), the model with bidirectional recurrence is depicted. We show in (d) how we added the temporal convolutions and three-dimensional max-pooling (MP refers to max-pooling). Panel (e) illustrates the architecture with added temporal convolutions and bidirectional recurrence.}
\label{fig:arch}
\end{figure}

\section{Architectures} \label{sec:approach}

% All our models have a 
% An overview of the different architectures is depicted in Figure \ref{fig:arch}.
In this section, we briefly describe the different architectures we investigate for gesture recognition in video. An overview of the models is depicted in Figure \ref{fig:arch}. Note that we pay close attention to the comparability of the network structures. The number of units in the fully connected layers and the number of cells in the recurrent models are optimized based on validation results for each network individually. All other hyper-parameters mentioned in this section and in Section \ref{sec:train} are optimized for the temporal pooling architecture. As a result, improvements over our baseline models are caused by architectural differences rather than better optimization, other hyper-parameters or preprocessing. 

\subsection{Baseline Models}

% Our two baseline models are: a single-frame model and a network integrating a temporal feature pooling layer. 
% The single-frame architecture worked well for \cite{karpathy2014large}, but is not a very fitting solution for our frame-wise gesture recognition setting. Nevertheless, this will give us an indication on how much still images contribute to the recognition. 
% A big advantage of the temporal pooling model is that features are aggregated from multiple frames, producing learned spatiotemporal features.

\textbf{Single-Frame}\, The single-frame architecture (Figure \ref{fig:archsingle}) worked well for general video classification \citep{karpathy2014large}, but is not a very fitting solution for our frame-wise gesture recognition setting. Nevertheless, this will give us an indication on how much static images contribute to the recognition. It has $3$$\times$$3$ convolution kernels in every layer. Two convolutional layers are stacked before performing max-pooling on non-overlapping $2$$\times$$2$ spatial regions. The shorthand notation of the full architecture is as follows: $C(16)$ - $C(16)$ - $P$ - $C(32)$ - $C(32)$ - $P$ - $C(64)$ - $C(64)$ - $P$ - $C(128)$ - $C(128)$ - $P$ - $D(2048)$ - $D(2048)$ - $S$, where $C(n_c)$ denotes a convolutional layer with $n_c$ feature maps, $P$ a max-pooling layer, $D(n_d)$ a fully connected layer with $n_d$ units and $S$ a softmax classifier. We deploy leaky rectified linear units (leaky ReLUs) in every layer. Their activation function is defined as $ a: x \mapsto \max(\alpha x, x)$, where $\alpha=0.3$. Leaky ReLUs seemed to work better than conventional ReLUs and showed promising results in other work \citep{maas2013rectifier,graham2014spatially,ndsb,xu2015empirical}. 

% The activation function of every unit in our model is the leaky rectified linear nonlinearity (LReLU) \cite{maas2013rectifier} defined as $ a: x \mapsto \max(\alpha x, x)$,
% where we choose $\alpha=0.3$.  This seemed to work better than
% \[ a(x) =
%   \begin{cases}
%     x       & \quad \text{if } x > 0\\
%     \alpha x  & \quad \text{otherwise}\\
%   \end{cases}
% \]

\textbf{Temporal Feature Pooling} \, The second baseline model exploits a temporal feature pooling strategy. As suggested by \cite{ng2015beyond}, we position the temporal pooling layer right before the first fully connected layer as illustrated in Figure \ref{fig:archpool}. This layer performs either mean-pooling or max-pooling across all video frames. The structure of the CNN-component is identical to the single-frame model. This network is able to collect all the spatial features in a given time window. However, the order of the temporal events is lost due to the nature of pooling across frames.

% \subsection{Single-Frame}

\subsection{Bidirectional Recurrent Models} \label{sec:rnn}

The core idea of RNNs is to create internal memory to learn the temporal dynamics in sequential data. 
% They have a recurrent character, because temporally unfolding the network reveals their hidden connections through time. 
An issue (in our case) with conventional recurrent networks is that their states are built up from \emph{previous} time steps. A gesture, however, generally becomes recognizable only after a few time steps, while the frame-wise nature of the problem requires predictions from the very first frame. This is why we use \emph{bidirectional} recurrence, which enables us to process sequences in both temporal directions. 

Describing the proposed model (Figure \ref{fig:archrec}) formally, we start with the CNN (identical to the single-frame model) transforming an input frame $x_t$ to a more compact vector representation $v_t$:
\begin{align}
v_t &= \textnormal{CNN}(x_t).
\end{align}
A bidirectional RNN computes two hidden sequences: the forward hidden sequence $h^{(f)}$ and the backward hidden sequence $h^{(b)}$:
% \begin{align}
% v_t &= \textnormal{CNN}(x_t) \\
% h_t^{f} &= a(W_{v}v_t+W_{f}h_{t-1}^{f}+b_f)\\
% h_t^{b} &= a(W_{v}v_t+W_{b}h_{t-1}^{b}+b_b)\\
% y_t &= \textnormal{softmax}(W_{d}(h_{t}^{f}+h_{t}^{b})+b_y)\\
% \end{align}
%
% Bidirectional recurrence:
\begin{align}
h_t^{(f)} &= \mathcal{H}_{f}(v_t, h_{t-1}^{(f)}) \quad \textnormal{and}\\
h_t^{(b)} &= \mathcal{H}_{b}(v_t, h_{t+1}^{(b)}),
% h_t^{b} &= a(W_{v}v_t+W_{b}h_{t-1}^{b}+b_b)\\
\end{align}
where $\mathcal{H}$ represents a recurrent layer and depends on the type of memory cell. There are two different cell types in widespread use: standard cells and LSTM cells \citep{hochreiter1997long} (we use the modern LSTM cell structure with peephole connections \citep{gers2003learning}). Both cell types will be compared in this work.
% Standard cells weight the input vector $v_t$ with trainable parameters $W_{vh}$ and summates with the previous hidden units $h_{t-1}$, weighted by $W_{vh}$, and a bias $b_h$.
% Standard cells are defined by
% %
% \begin{align}
%  h_t &= a(W_{vh}v_t+W_{hh}h_{t-1}+b_h),
% \end{align}
% where $W_{vh}$, $W_{hh}$ and $b_h$ are trainable parameters and $a$ is the same leaky rectified linear nonlinearity as used in the CNN. LSTMs cells are more complex, but their structure allows them to hold memory for much longer, hence the name. This enables them to capture long-range temporal dependencies. The cells can be described as follows:
% \begin{align}
% i_t &= \sigma(W_{vi}v_t+W_{hi}h_{t-1}+w_{ci}\odot c_{t-1}+b_i),\\
% f_t &= \sigma(W_{vf}v_t+W_{hf}h_{t-1}+w_{cf}\odot c_{t-1}+b_f),\\
% o_t &= \sigma(W_{vo}v_t+W_{ho}h_{t-1}+w_{co}\odot c_{t-1}+b_o),\\
% g_t &= \tanh(W_{vg}v_t+W_{hg}h_{t-1}+b_g),\\
% c_t &= f_t \odot c_{t-1} +  i_t \odot g_t,\\
% h_t &= o_t \odot \tanh(c_t),
% \end{align}
% where $\odot$ denotes the point-wise multiplication of two vectors and all parameters referred by $W_.$, $w_.$ or $b_.$ are trainable.
Finally, the output predictions $y_t$ are computed with a softmax classifier which takes the sum of the forward and backward hidden states as input:
\begin{align}
y_t &= \textnormal{softmax}(W_{y}(h_{t}^{(f)}+h_{t}^{(b)})+b_y).
\end{align}
% The proposed model is illustrated in Figure \ref{fig:archrec}.

% \begin{figure}[tb]
% \begin{center}
% \begin{subfigure}{0.49\textwidth}
% \centering
% \resizebox{!}{3cm}{\input{imgs/single.tex}}
% \caption{Single-Frame}
% \label{fig:archsingle}
% \end{subfigure}
% %
% \begin{subfigure}{0.49\textwidth}
% \centering
% \resizebox{!}{3cm}{\input{imgs/pool.tex}}
% \caption{Temp Pooling}
% \label{fig:archpool}
% \end{subfigure}
% %
% \end{center}
% \caption{\textbf{Baseline models.} Panel (a) shows a single-frame CNN architecture. Panel (b) illustrates a temporal feature pooling network (max- or mean-pooling) proposed by \cite{ng2015beyond} spanning multiple video frames. The proposed models are illustrated in Figure \ref{fig:arch}.}
% \label{fig:baseline}
% \end{figure}

\subsection{Adding Temporal Convolutions} \label{sec:3d}

% \begin{align}
% s_{tij}^{(k)} &=  
%        \sum_{n=1}^{N} 
%                     \sum_{t'=1}^{T} 
%             W^{(kn)}_{t'} x^{(n)}_{t-t',i,j} \\
% z_{tij}^{(k)} &=  
%         b^{(k)} + \sum_{m=1}^{M}
%                     \sum_{i'=1}^{I} 
%                     \sum_{j'=1}^{J} 
%             W^{(kn)}_{i'j'} s^{(m)}_{t,i-i',j-j'} 
%             \\
% v_{ij}^{(k)} &= a(z_{tij}^{(k)})
% \end{align}

Our final set of architectures extends the CNN layers with temporal convolutions (convolutions over time). This enables the extraction of hierarchies of motion features and thus the capturing of temporal information from the first layer, instead of depending on higher layers to form spatiotemporal features. Performing three-dimensional convolutions is one approach to achieve this. However, this leads to a significant increase in the number of parameters in every layer, making this method more prone to overfitting. Therefore, we decide to factorize this operation into two-dimensional spatial convolutions and one-dimensional temporal convolutions. This leads to fewer parameters and optionally more nonlinearity if one decides to activate both operations. We opt to not include a bias or another nonlinearity in the spatial convolution step to maintain the comparability between architectures.

% To alleviate confusion early on, we aren't performing true three-dimensional convolutions. We'll explain later why. 

% First, we compute spatial feature maps $s_t$ for every frame $x_t$. A pixel on position $(i,j)$ of the $k$-th feature map is determined as follows:
% %
% \begin{align}
% s_{tij}^{(k)} &=  
%        				\sum_{n=1}^{N}
%                     \sum_{p=1}^{P} 
%                     \sum_{q=1}^{Q} 
%             W^{(kn)}_{pq} x^{(n)}_{t,i-p,j-q}\,,
% \end{align}
% where $N$ is the number of input channels with dimensions $P$$\times$$Q$. Finally, we convolute across the time dimension for every position $(i,j)$, summate the bias $b^{(k)}$ and output the activation function $a$:
% %
% \begin{align}
% v_{tij}^{(k)} &= a \left( 
%         b^{(k)} + \sum_{m=1}^{M} 
%                     \sum_{t'=1}^{T} 
%             W^{(km)}_{t'} s^{(m)}_{t-t',i,j}  \right),
% \end{align}
% where variables referred by $W_.$, or $b_.$ are trainable parameters.

First, we compute spatial feature maps $s_t$ for every frame $x_t$. A pixel at position $(i,j)$ of the $k$-th feature map is determined as follows:
\begin{align}
s_{tij}^{(k)} &=  
       				\sum_{n=1}^{N}
            \left(W_{\textnormal{spat}}^{(kn)} * x^{(n)}_{t}\right)_{ij}\,,
\end{align}
where $N$ is the number of input channels and $W_{\textnormal{spat}}$ are trainable parameters. Finally, we convolve across the time dimension for every position $(i,j)$, add the bias $b^{(k)}$ and apply the activation function $a$:
\begin{align}
v_{tij}^{(k)} &= a \left( 
        b^{(k)} + \sum_{m=1}^{M} 
                    % \sum_{t'=1}^{T} 
            \left(W^{(km)}_{\textnormal{temp}} * s^{(m)}_{ij}\right)_{t}  \right),
\end{align}
where the variables $W_{\textnormal{temp}}$ and $b$ are trainable parameters and $M$ is the number of spatial feature maps.

% To put the above formulation in other words, we ``insert'' one-dimensional convolutions to the two-dimensional convolutional layers. Performing three-dimensional convolutions instead, would drastically increase the number of parameters. For example, computing our convolutional layer on $16$ input channels to form $16$ feature maps with $3$$\times$$3$-sized spatial kernels and temporal kernels of size $3$, result in $3072$ parameters. Applying $3$$\times$$3$$\times$$3$ filters result in $6912$ parameters, which is more than twice as much. To this end, we are able to learn motion features throughout the network while avoiding an explosion in number of parameters.

% \textbf{Adding Temporal Convolutions.} 
Two different architectures are proposed using this new layer. In the first model (Figure \ref{fig:arch3d}), we replace the convolutional layers of the single-frame CNN with the spatiotemporal layer defined above. Furthermore, we apply three-dimensional max-pooling to reduce spatial as well as temporal dimensions while introducing slight translational invariance in time. Note that this architecture implies a sliding window approach for frame-wise classification, which is computationally intensive.
%
% \textbf{Adding Temporal Convolutions and Recurrence.} 
In the second model, illustrated in Figure \ref{fig:arch3drec}, the time dimensionality is retained throughout the network. That means we only carry out spatial max-pooling. To this end, we are able to stack a bidirectional RNN with LSTM cells, responding to high-level temporal dependencies. It also incidentally resolves the need for a sliding window approach to implement frame-wise video classification.

% \begin{align}
% s_{ijt}^{k} &=  (W_m^k *_t x)_t + b_m^k)\\
% h_{ijt}^{k} &=  a \left( (W_s^k *_{ij} s)_{ij} + b_s^k \right) \\
% \end{align}

\section{Experiments} \label{sec:exp}

\subsection{Montalbano Gesture Recognition Dataset} \label{sec:dataset}
The ChaLearn Looking At People (LAP) 2014 Challenge \citep{chalearn14} consists of three tracks: human pose recovery, human action/interaction recognition and gesture recognition. The dataset accompanying the gesture recognition challenge, called the Montalbano dataset, will be used throughout this work. The dataset is multi-modal, because the gestures are captured with a Microsoft Kinect that has a depth sensor. In all sequences, a single user is recorded in front of the camera, performing natural communicative Italian gestures. Each data file contains an RGB-D (where ``D'' stands for depth) image sequence and a skeletal pose stream provided by the Microsoft Kinect API. The gesture vocabulary contains $20$ Italian cultural/anthropological signs. The gestures are not segmented, which means that sequences typically contain several gestures. Gesture performances appear randomly within the sequence without a prearranged rest pose. Moreover, several unannotated out-of-vocabulary gestures are present. 

It is the largest publicly available gesture dataset of its kind. There are $1,720,800$ labeled frames across $13,858$ video fragments of about $1$ to $2$ minutes sampled at $20$Hz with a resolution of $640$$\times$$480$. The gestures are performed by $27$ different individuals under diverse conditions; these include varying clothes, positions, backgrounds and lighting. The training set contains $11,116$ gestures and the test set contains $2,742$. The class imbalance is negligible. The starting and ending frames for each gesture are annotated as well as the gesture class label.

To speed up the training, we crop part of the images containing the user and rescale them to $64$ by $64$ pixels using the skeleton information (other than that, we do not use any pose data).
% The only preprocessing we perform is cropping the images to where the user is positioned and rescaling them to $64$ by $64$ pixels. This was easily performed by using the provided skeleton information. If no depth-sensing camera would be available, this could easily be done with face tracking technology. 
However, we show in Section \ref{sec:results} that we even achieve good results when we do not crop the images and leave out depth information.

\subsection{End-To-End Training} \label{sec:train}

We train our models from scratch in an end-to-end fashion, backpropagating through time (BTT) for our recurrent architectures. The network parameters are optimized by minimizing the cross-entropy loss function using mini-batch gradient descent with the \emph{Adam} update rule \citep{kingma2014adam}. 
% Adam is a recently introduced optimization algorithm based on adaptive estimates of lower-order moments of the gradients. 
We found that Adam works great in practice, especially when experimenting with very different layer types in the same model.  
% Leaving the proposed hyper-parameters of Adam untouched, we observed improved training convergence in comparison to SGD with Nesterov momentum. 
All our models are trained the same way with early stopping, a mini-batch size of $32$, a learning rate of $10^{-3}$ and an exponential learning rate decay. 
% To cope with memory constraints, we used micro-batches where the gradients are summated cumulatively. Once a mini-batch size is reached, the gradients are divided by the number of micro-batches performed and passed to the Adam algorithm.
Before training, we initialize the weights with a random orthogonal initialization method \citep{saxe2013exact}.
 % One exception are the weights of LSTM cells, which are initialized using a normal distribution with $\mu=0$ and $\sigma=10^{-2}$.

\textbf{Recurrent Networks}\, As described in Section \ref{sec:dataset}, the video files in the Montalbano dataset contain approximately 1 to 2 minutes of footage, consisting of multiple gestures. Recurrent models are trained on random fragments of $64$ frames and produce $64$ predictions, one for every frame. To summarize, a data sample has $4$ channels (RGB-D), $64$ frames each, with a resolution of $64$ by $64$ pixels; or in shorthand notation: $4@64$$\times$$64$$\times$$64$. We optimized the number of cells for each model based on validation results. For LSTM cells, we only saw a small improvement between $512$ and $1024$ units, so we settled at $512$. For RNNs with standard cells, we used $2048$ units. 
The location of gestures within the long sequences is not given. A gesture is generally about 20 to 50 frames long. If a small fraction of a gesture is located at the beginning or the end of the $64$ considered frames, the model does not have enough information to label these frames correctly. That is why we allow a buildup in both forward and backward direction for evaluation; we feed $64$ frames into the RNN and keep the middle $32$ for evaluation.
% To compare, this is nearly $7$ times more input pixels than used by Krizhevsky et al. \cite{krizhevsky2012imagenet} for ImageNet. 

\textbf{Non-Recurrent Networks}\, The single-frame CNN is trained frame by frame and all other non-recurrent networks are trained with the number of frames optimized for their specific architecture. The best number of frames to mean-pool across is $32$, determined by validation scores with tested values in $[8,16,32,64]$. In the case of max-pooling, we find that pooling over $16$ frames gives better outcomes. Also, pretraining the CNNs frame-by-frame and fine-tuning with temporal max-pooling gave slightly improved results. We observed no improvements, however, using this technique with temporal mean-pooling. The architecture with added temporal convolutions and three-dimensional max-pooling showed optimal results by considering $32$ surrounding frames. The targets for all the non-recurrent networks are the labels associated with the centermost frame of the input video fragment. We evaluate these models using a sliding window with single-frame steps.

\textbf{Regularization and Data-Augmentation}\, We employed many different methods to regularize the deep networks. Data augmentation has a significant impact on generalization. For all our trained models, we used the same augmentation parameters: $[-5,5]$ pixel translations in vertical direction and $[-10,10]$ horizontal, $[-2,2]$ rotation degrees, $[-2,2]$ shearing degrees, $[\frac{1}{1.1},1.1]$ image scaling factors and $[\frac{1}{1.2},1.2]$ temporal scaling factors. From each of these intervals, we sample a random value for each video fragment and apply the transformations online using the CPU. Dropout with $p=0.5$ is used on the inputs of every fully connected layer. Furthermore, using leaky ReLUs instead of conventional ReLUs and factorizing three-dimensional convolutions into spatial and temporal convolutions also reduce overfitting.

\begin{table}[t]
\centering
\begin{tabular}{lcccc}
\toprule
\multicolumn{1}{l}{\bf Architecture}  &\multicolumn{1}{c}{\bf Jaccard Index}  &\multicolumn{1}{c}{\bf Precision} &\multicolumn{1}{c}{\bf Recall} &\multicolumn{1}{c}{\bf Error Rate*}\\
\midrule

Single-Frame CNN (Figure \ref{fig:archsingle})					&0.465 &67.86\% &57.57\% &20.68\%\\
Temp Max-Pooling (Figure \ref{fig:archpool})					&0.748 &85.03\% &82.92\% &8.66\%\\
Temp Mean-Pooling (Figure \ref{fig:archpool})					&0.775 &85.93\% &85.80\% &8.55\%\\
Temp Conv (Figure \ref{fig:arch3d})							&0.842 &89.36\% &90.15\% &4.67\%\\
RNN, Standard Cells (Figure \ref{fig:archrec})				&0.885 &92.77\% &93.56\% &3.58\%\\
RNN, LSTM Cells (Figure \ref{fig:archrec})					&0.888 &93.75\% &93.28\% &3.55\%\\
Temp Conv + LSTM (Figure \ref{fig:arch3drec})					&\textbf{0.906} &\textbf{94.49\%} &\textbf{94.57\%} &\textbf{2.77\%}\\
\bottomrule
\end{tabular}
\caption{A comparison of the results for our different architectures on the Montalbano gesture recognition dataset. The Jaccard index indicates the mean overlap between the binary predictions and the binary ground truth across gesture categories. We also compute precision and recall scores for each gesture class and report the mean score across classes. \\ *The error rate is based on majority voted frame-wise predictions from \emph{isolated} gesture fragments. }
\label{tab:results}
\end{table}

\begin{table}[bt]
\begin{center}
\begin{tabular}{lcccc}
\toprule
\multicolumn{1}{l}{\bf Model} &\multicolumn{1}{c}{\bf Crop} &\multicolumn{1}{c}{\bf Depth} &\multicolumn{1}{c}{\bf Pose} &\multicolumn{1}{c}{\bf Jaccard Index} \\ 
\midrule
\cite{chang2014nonparametric} (MRF, KNN, PCA, HoG) &\textbf{yes} &no &\textbf{yes} 	&0.827 \\
\cite{monnier2014multi} (AdaBoost, HoG) &\textbf{yes} &\textbf{yes} &\textbf{yes} 	&0.834 \\
\cite{neverova2014moddrop} (Multi-Scale DNN)  &\textbf{yes} &\textbf{yes} &no 	&0.836 \\
\cite{neverova2014moddrop} (Multi-Scale DNN)  &\textbf{yes} &\textbf{yes} &\textbf{yes} 	&0.870 \\
% Peng et al. \cite{neverova2014moddrop} &\textbf{yes} &\textbf{yes} &\textbf{yes} &no	&87.0\% \\
% Pigou et al. \cite{pigou2014sign} &\textbf{yes} &\textbf{yes} &\textbf{yes} &no	&87.0\% \\
\midrule 
 										&no &no &no 	&0.842 \\
Temp Conv + LSTM 										&\textbf{yes} &no &no 	&0.876 \\
  										&\textbf{yes} &\textbf{yes} &no 	&\textbf{0.906} \\	
% \midrule 
% \midrule 
\bottomrule
\end{tabular}
\end{center}
\caption{Montalbano gesture recognition dataset results compared to previous work. \emph{Crop}: the cropping of specific areas in the video using the skeletal information. \emph{Depth}: the usage of depth-maps. \emph{Pose}: the usage of the skeletal stream as features. 
% \emph{Audio}: a spoken component where the users say the gesture out loud.
Note that even when we do not use depth images, we still achieve better results.\\}
\label{tab:sota}
\end{table}

\subsection{Results} \label{sec:results}

We follow the ChaLearn LAP 2014 Challenge score to measure the performance of our architectures. This way, we can compare with previous work on the Montalbano dataset. The competition score is based on the Jaccard index, which is defined as follows:
\begin{align}
J_{s,n} &= \frac{|A_{s,n} \cap B_{s,n} |}{|A_{s,n} \cup B_{s,n}|}.
\end{align}
The binary ground truth for gesture category $n$ in sequence $s$ is denoted as the binary vector $A_{s,n}$, whereas $B_{s,n}$ denotes the binary predictions. The Jaccard index $J_{s,n}$ can be seen as the overlap rate between $A_{s,n}$ and $B_{s,n}$. To compute the final score, the mean Jaccard index among all categories and sequences is computed:
\begin{align}
J_{\textnormal{avg}} &= \frac{1}{N S}\sum_{s=1}^{S} \sum_{n=1}^{N} J_{s,n},
\end{align}
where $N=20$ is the number of categories and $S$ the number of sequences in the test set. 
% To provide a more profound indication of classification performance, we also consider precision and recall measures. We do this by computing precision and recall for each class individually followed by aggregating the mean.

\begin{figure}[t]
\begin{center}
\includegraphics[width=\textwidth]{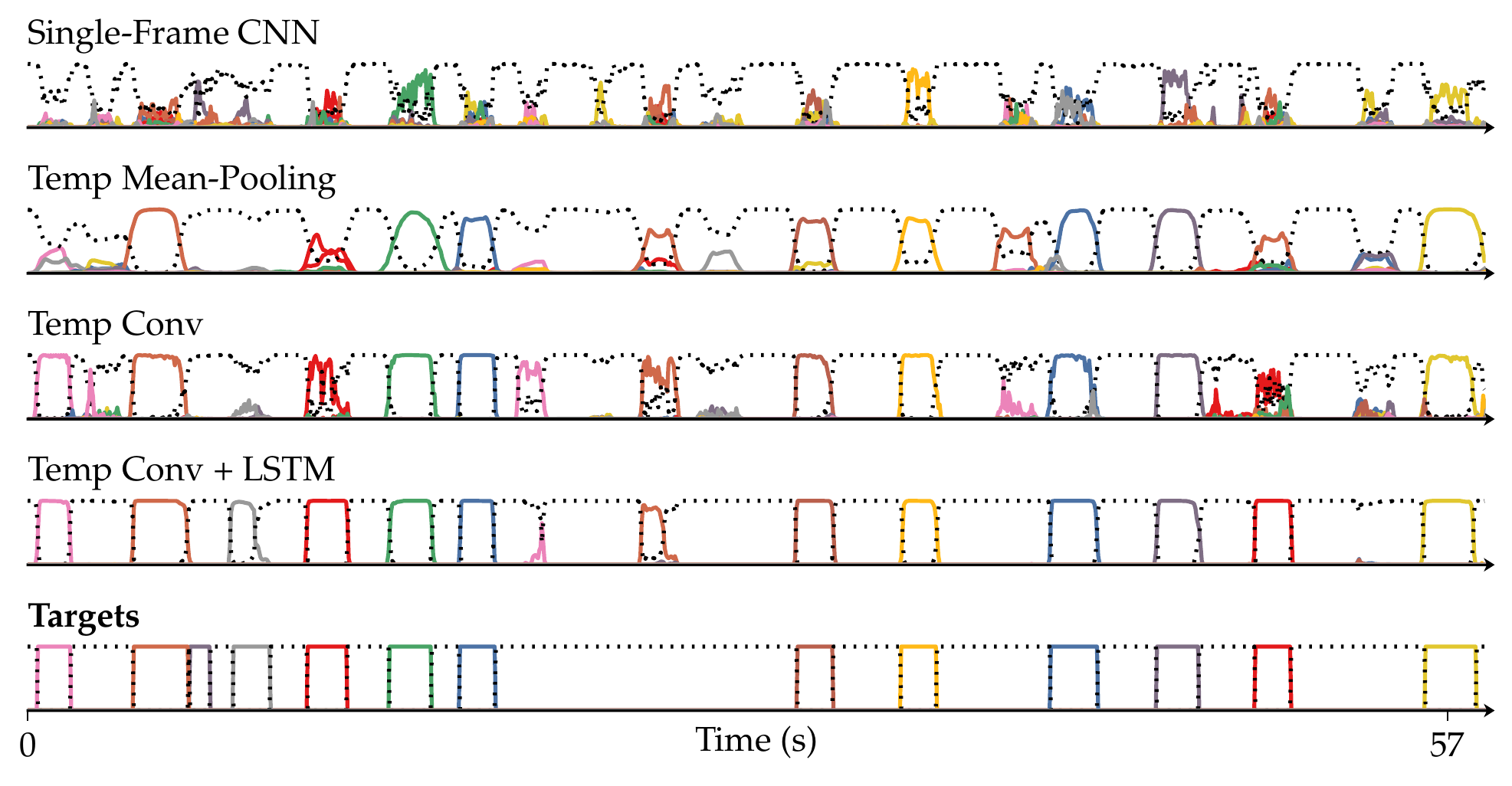}
\end{center}
\caption{The output probabilities are shown for a sequence fragment in the test set. The dashed line represents silences. The non-recurrent models make more mistakes and have difficulties making hard decisions to where the gesture starts or ends and are unable to smooth out predictions in time. Adding recurrence enables deep networks to learn the behavior of the manual annotators with great accuracy.}
\label{fig:classprob}
\end{figure}

An overview of the results for our different architectures is shown in Table \ref{tab:results}. The predictions of the single-frame baseline achieve a Jaccard index below $0.5$. This is to be expected as no motion features are extracted. We observe a significant improvement with temporal feature pooling (a Jaccard index of $0.775$ vs. $0.465$). 
Furthermore, mean-pooling performs better than max-pooling. 
% Mean-pooling treats the considered frames equally, while max-pooling takes the maximum activation for each feature across frames. This might confuse the model at gesture boundaries, because it is unpredictable which frames cause these strong activations. 
Adding temporal convolutions and three-dimensional max-pooling improves the Jaccard index to $0.842$.

The three last entries in Table \ref{tab:results} use recurrent networks. Surprisingly, the RNNs are only acting on high-level spatial features, yet are surpassing a CNN learning hierarchies of motion features (a Jaccard index of $0.842$ vs. $0.888$). The difference in performance for the two types of cells is very small and they can be considered equally capable for this type of problem where temporal dependencies are not too long-ranged. Finally, combining the temporal convolution architecture with an RNN using LSTM cells improves the score even more ($0.906$). This deep network not only learns multi-level spatiotemporal features, but is capable of modeling temporal dynamics within them.

In Table \ref{tab:sota}, we compare our results with previous work.  Our best model outperforms the method of \cite{neverova2014moddrop} when we only consider RGB-D pixels as input features ($0.906$ vs. $0.836$). When we remove depth information and perform no preprocessing other than rescaling the images, we still achieve better results ($0.842$). 
Moreover, we even achieve better results without the need for depth images or pose information ($0.876$ vs. $0.870$). 
% Their results with audio features (the users say the gesture out loud) is a Jaccard index of $0.880$.

% , even when we pretend the footage was filmed with a standard RGB camera and perform no preprocessing other than rescaling the images ($83.6\%$ vs. $84.2\%$ overlap).

To illustrate the differences in output predictions of the different architectures, we show them for a randomly selected sequence in Figure \ref{fig:classprob}. We see that the single-frame CNN has trouble classifying the gestures, while the temporal pooling is significantly more accurate. However, the latter still has difficulties with boundaries. Adding temporal convolutions shows improved results, but the output contains more jagged predictions. This seems to disappear by introducing recurrence. The output of the bidirectional RNN matches the target labels strikingly well. 
% The temporal pooling model, however, has difficulties with boundaries and makes more mistakes in general.

\begin{figure}[tb]
\def \figspace {0.32\textwidth}
\def \rsize {2cm}
\captionsetup[subfigure]{labelformat=empty}
% \begin{center}
% \hspace*{\fill}%
\captionsetup[subfigure]{justification=justified,singlelinecheck=false}
\begin{subfigure}{0.48\textwidth}
	% \centering
	\caption{\textbf{While standing still}}
	\captionsetup[subfigure]{justification=centering,singlelinecheck=true, labelformat=empty}
	\setcounter{subfigure}{3}
	\begin{subfigure}[t]{\figspace}
	% \centering
	\resizebox{\rsize}{!}{\includegraphics[width=\textwidth]{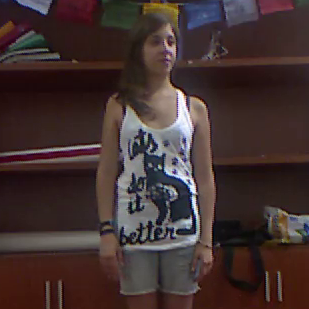}}
	\caption{RGB Input}
	\label{fig:srgb}
	\end{subfigure}
	\begin{subfigure}[t]{\figspace}
	\centering
	\resizebox{\rsize}{!}{\includegraphics[width=\textwidth]{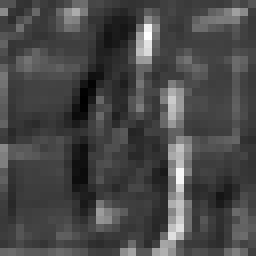}}
	\caption{Spatial\\Feature Map}
	\label{fig:ss}
	\end{subfigure}
	\begin{subfigure}[t]{\figspace}
	\centering
	\resizebox{2.1cm}{!}{\includegraphics[width=\textwidth, cframe=red]{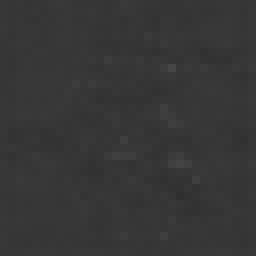}}
	\caption{\textbf{Spatiotemporal\\Feature Map}}
	\label{fig:st}
	\end{subfigure}
\end{subfigure}
%
% \unskip\ \vrule\
\hfill
\begin{subfigure}{0.48\textwidth}
	\caption{\textbf{While moving}}
	\captionsetup[subfigure]{justification=centering,singlelinecheck=true, labelformat=empty}
	\setcounter{subfigure}{0}
	% \centering
	\begin{subfigure}[t]{\figspace}
	\centering
	\resizebox{2.07cm}{!}{\input{imgs/movergb.tex}}
	\caption{RGB Input}
	\label{fig:mrgb}
	\end{subfigure}
	\begin{subfigure}[t]{\figspace}
	\centering
	\resizebox{\rsize}{!}{\includegraphics[width=\textwidth]{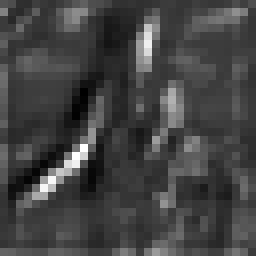}}
	\caption{Spatial\\Feature Map}
	\label{fig:ms}
	\end{subfigure}
	\begin{subfigure}[t]{\figspace}
	\centering
	\resizebox{2.1cm}{!}{\includegraphics[width=\textwidth, cframe=red]{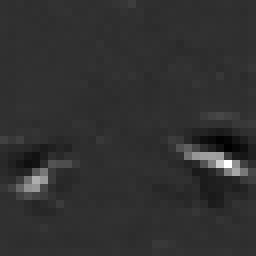}}
	\caption{\textbf{Spatiotemporal\\Feature Map}}
	\label{fig:mt}
	\end{subfigure}
\end{subfigure}
% \hspace*{\fill}%
% \hfill
%
% \end{center}
\caption{\textbf{Motion Features}\, This figure illustrates the effect of integrating temporal convolutions. The depicted spatial feature map is the most active 4-layer-deep feature map, extracted from an architecture without temporal convolutions. The spatiotemporal feature map is extracted from a model with temporal convolutions. The strong activations in the spatiotemporal feature maps while moving indicate learned motion features.}
\label{fig:motion}
\end{figure}

In Figure \ref{fig:motion}, we show that adding temporal convolutions enables neural networks to capture motion information. When the user is standing still, the units of the feature map are inactive, while the feature map from the network without temporal convolutions has a lot of active units. When the user is moving, the feature map shows strong activations at the movement locations. This suggests that the model has learned to extract motion features.

\fboxrule=2pt%border thickness

\section{Conclusion and Future Work} \label{sec:conclusion}
% With our baseline models being a single-frame architecture and a temporal feature pooling network, 
We showed in this paper that adding bidirectional recurrence and temporal convolutions improves frame-wise gesture recognition in video significantly. We observed that RNNs responding to high-level spatial features perform much better than single-frame and temporal pooling architectures, without the need to take into account the temporal aspect in the lower layers of the network. However, adding temporal convolutions in all layers of the architecture has a notable impact on the performance, as they are able to learn hierarchies of motion features, unlike RNNs. Standard cells and LSTM cells appear to be equally strong for this problem. 
% This suggests that the added complexity of LSTMs is not required for gesture recognition. 
Furthermore, we observed that RNNs outperform non-recurrent networks and are able to predict the beginning and ending frames of gestures with great accuracy, whereas other models show uncertainty at these boundaries.

% \section{Future Work} \label{sec:fw}

In the future, we would like to build upon this work for research in the domain of sign language recognition.
This is even more challenging than gesture recognition. The vocabulary is larger, the differences in finger positions and hand movements are more subtle and signs are context dependent, as they are part of a language. Sign language is not related to written or spoken language, which complicates annotation and translation. Moreover, signers communicate simultaneously with facial, manual (both hands are separate communication channels) and body expressions. This means that sign language video cannot be translated the way speech recognition can transcribe audio to written sentences.
% In the future, we would like to experiment with a dataset, whether or not with artificial data, that would achieve an overlap of close to $0\%$ with a single-frame architecture. To this end, the ability of a deep network to learn motion features would be even more apparent from the scores.

\subsubsection*{Acknowledgments}
We would like to thank NVIDIA Corporation for the donation of a GPU used for this research.
The research leading to these results has received funding from the Agency for Innovation by Science and Technology in Flanders (IWT).

% \clearpage
\subsubsection*{References}
\renewcommand{\section}[2]{}%
% \nocite{*}r
\bibliography{lib}
\bibliographystyle{iclr2016_conference}

\end{document}